\documentclass{article}
\pdfoutput=1
\usepackage[final]{nips_2017}
\usepackage[utf8]{inputenc} % allow utf-8 input
\usepackage[T1]{fontenc}    % use 8-bit T1 fonts
\usepackage{hyperref}       % hyperlinks
\usepackage{booktabs}       % professional-quality tables
\usepackage{amsfonts}       % blackboard math symbols
\usepackage{nicefrac}       % compact symbols for 1/2, etc.
\usepackage{microtype}      % microtypography
\usepackage{graphicx}

\usepackage{float}
\usepackage{amsfonts}
\usepackage{booktabs}
\usepackage{siunitx}
\usepackage{xurl}
\usepackage{subcaption}
\usepackage{bm}

\title{Face Verification Bypass}

\author{
  Sanjana Sarda \\
  Department of Electrical Engineering\\
  Stanford University\\
  \texttt{ssarda@stanford.edu} \\
}

\begin{document}
\maketitle

\begin{abstract}
    Face verification systems aim to validate the claimed identity using feature vectors and distance metrics. However, no attempt has been made to bypass such a system using generated images that are constrained by the same feature vectors. In this work, we train StarGAN v2 to generate diverse images based on a human user that have similar feature vectors yet qualitatively look different. We then demonstrate a proof of concept on a custom face verification system and verify our claims by demonstrating the same proof of concept in a black box setting on dating applications that utilize similar face verification systems. 
\end{abstract}

\section{Introduction}
Face verification unlike identification, is concerned with validating a claimed identity based on the image of a face by either accepting or rejecting the identity claim (one to one matching). Several popular applications have started employing face verification techniques to guarantee user identity. For example, dating applications such as Bumble and Tinder utilize photo verification to guarantee that a user on the platform is who they claim to be. This works by asking the user to take a picture using the in-app camera which is then compared with the pictures that the user has on their profile. 

Considering this methodology, it provokes the question if it is possible to essentially create a fake profile using photos that can be verified with the user's face. It is important to note that for this concept to work, the generated face must have similar enough features such that it can pass a face verification system while qualitatively looking different. 

In this work, we demonstrate that it is possible to generate images that have similar feature vectors or embeddings which can be used to bypass face verification systems.

\section{Related Work}
\textbf{Attacking the face recognition authentication}\cite{Securing}. This blog post produces novel research to attack facial recognition systems using various attack vectors such as classic API vulnerabilities and more interestingly, biometric vulnerabilities. Some naive implementations involve checking if an uploaded video was performed by the same person, which is easy to bypass. However, for more advanced implementations, the person in the video is correlated with the original photo used for identification. The authors used an open-source “faceswap” project to generate fake recordings of the victim to attack this specific feature, however, they did not report any results.

\noindent \textbf{DeepPrivacy}\cite{hukkelaas2019deepprivacy} The authors of DeepPrivacy leverage a generative adversarial network to automatically anonymize faces in images to bypass facial recognition systems by replacing the original face with a realistic generated face. The face is randomly generated based on a dataset of human faces consisting of different poses and backgrounds. 

\noindent \textbf{Face Recognition Review}\cite{anwarul2020comprehensive}. This paper consists of a literature review of the factors that facial recognition tend to be dependent on such as aging, pose, variation, partial occlusion, illumination, and facial expressions as well as the techniques that are used to mitigate issues caused by these factors. Face recognition classification methods can either be appearance-based, feature-based, or a hybrid of the two. Recent approaches also involve deep reinforcement learning with CNNs.

\noindent \textbf{Master Face Attacks on Face Recognition Systems}\cite{nguyen2021master} \& \noindent \textbf{Master Faces for Dictionary Attacks}\cite{shmelkin2021generating}. Recent work has suggested that it is possible to bypass facial verification systems with a single master face. The authors of \cite{shmelkin2021generating} used a greedy coverage search to find an image within the specific dataset that consisted of the most similar features. However, the master faces have been generated using biased training data (LFW) and hence are not very accurate (only successful for ~40\% of test data) and are specifically male Caucasians with white hair. Besides this, the authors have not tested their master face on real-world systems and data. It is also possible that a more accurate “master face” is not a face at all. It also may be beneficial to focus on a single verification system at a time such that it guarantees a higher accuracy score. 

\noindent \textbf{Black-Box Adversarial Attacks}\cite{xu2020adversarial}. This work is related to black-box adversarial attacks in that the bypass depends on how robust the verification system is to different amounts of feature perturbation.

At the time of writing, no solution has been proposed that guarantees the generation of dissimilar facial images to bypass facial verification systems.

\section{Datasets}
For this project we will be using two separate datasets for training the face generator and building the face verification system respectively. 

\subsection{Human User Dataset}
This dataset currently consists of 310 images of the human user's face over 4 years to model the real world image space (unique lighting, angles, age). In consideration of image and model safety the author's face is used in the dataset. After extracting the face subset using a Caffe-based face detector \cite{liu2016ssd}, each image is cropped to the same dimension. 

\subsection{FairFace}
FairFace is a race balanced dataset consisting of 108,501 face images \cite{karkkainenfairface}. Each image is treated as a unique user in the verification system. Like in the previous dataset, after extracting the face subset using a Caffe-based face detector \cite{liu2016ssd}, each image is cropped to the same dimension.

\section{Face Verification Model}
The Face Verification Model is based off a general implementation of FaceNet \cite{schroff2015facenet} and DeepFace \cite{taigman2014deepface} using a pre-trained ConvNet Inception model \cite{szegedy2015going} that encodes each input image into a 128-dimensional vector. Image vectors are compared using the triplet loss function where 

\[ J = \sum_{i = 1} ^{m} [||f(A^i) - f(P^i)||_2^2 - ||f(A^i) - f(N^i)||_2^2 +\alpha \]

and A is the anchor image, P is a positive image, N is a negative image, and $\alpha$ is the margin. 

The system uses a database built on face images from the train subset of FairFace. To pass the facial verification system, the distance induced by the Frobenius norm of an image is calculated against the intended user in the database. If the distance is under the threshold of 0.7, the image is considered to be the same person, otherwise the verification fails. 

\section{Approach}

\subsection{Baseline}
For the baseline model we used a naive approach of fine tuning the StyleGAN \cite{karras2019style} model on the human user dataset such that it would stochastically generate an image that would pass the face verification system while qualitatively appearing to be different from images in the training dataset. For this purpose, we used the approach in \cite{mo2020freeze} to freeze first four layers in the discriminator so that it would not overfit on the training data and create overly similar images (diverse outputs are crucial). As in the paper, exponential moving averages of the weights are kept for the generator for inference. Additionally, mixing regularization is used for fine tuning.

\subsection{StarGAN v2}
The results from the baseline model qualitatively look similar to images from the training dataset (less diverse) and were of lower resolution (low fidelity). To solve this problem and also handle training from the direction of seed images towards the intended face, we decided to use the StarGAN v2 \cite{choi2020starganv2}. While StarGAN v2 has a pre-trained model trained on Celeb-A for evaluation purposes, checkpoints for training are not available so we first pre-trained the model. To prevent overfitting, we used the validation set from the FairFace (not used in the Face Verification Model) for the train and validation dataset and trained for approximately 10 hours. To handle memory issues, we changed the batch size to 4 and the validation size to 8. To generate images, we experimented with using the train data as reference (the seed image) images with processed images from the human user dataset as the source. We also experimented with using processed images from the human user dataset as both the reference and the source, with less useful results. 

\section{Experiments}

\subsection{Face Verification Experiments}
To verify that an image from the human user dataset could be used to verify an arbitrary image in the face verification system we conduct an initial experiment with a verification model built on a subset of 1000 images. To optimize the search, an image from the human user dataset was tested against all IDs in the verification system. Images from IDs that led to successful verifications were then tested against the human user ID. 

\begin{figure}[ht]
        \centering
        \begin{subfigure}{0.15\textwidth}
        \centering
        \includegraphics[width= \linewidth]{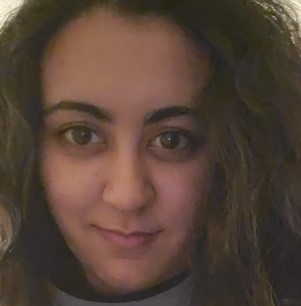}
         \caption{}
        \end{subfigure}\quad
        \begin{subfigure}{0.15\textwidth}
        \centering
        \includegraphics[width= \linewidth]{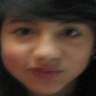}
        \caption{}
        \end{subfigure}\quad
        \begin{subfigure}{0.15\textwidth}
        \centering
        \includegraphics[width= \linewidth]{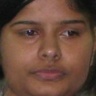}
         \caption{}
        \end{subfigure}\quad
        \caption{For the key image (extreme left), the image in the middle failed verification while the image on the extreme right passed verification. }
        \label{fig:1}
    \end{figure} 

It should be noted that the face verification model is dependent on the number of images it is built on. For a database built on face images from the entire train subset of FairFace, both (b) and (c) in Figure \ref{fig:1}, fail verification.
    
\subsection{Metrics}

\subsubsection{Distance based on Frobenius Norm (User Image to Test Image) - \textbf{FN}}
Since we are trying to find a test image to add to the Face Verification System that can be verified with the user's face, we observe the distance for a given test image with two different user images.

\subsubsection{Mahalanobis Distance (Test Image to Human User Dataset) - \textbf{MD}}
The  Mahalanobis distance between $\bm{x}^i$ and $\bm{x}^j$ is given by $\Delta^2=(\bm{x}^i-\bm{x}^j)^\top\Sigma^{-1}(\bm{x}^i-\bm{x}^j)$, where $\Sigma$ is a $d\times d$ covariance matrix. This metric is commonly used in image processing techniques for pattern and template matching. Since we are trying to find an image that qualitatively is the furthest away from the human user dataset, we observe the Mahalanobis distance between the feature vector of the test image with the feature vectors of images from the human user dataset.

\subsection{Baseline Experiments}
\begin{figure}[ht]
        \centering
        \begin{subfigure}{0.15\textwidth}
        \centering
        \includegraphics[width=\linewidth]{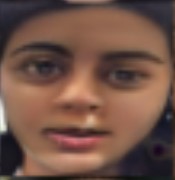}
        \caption{}
        \end{subfigure}\quad
        \begin{subfigure}{0.15\textwidth}
        \centering
        \includegraphics[width=\linewidth]{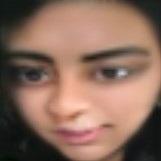}
        \caption{}
        \end{subfigure}\quad
        \begin{subfigure}{0.15\textwidth}
        \centering
        \includegraphics[width=\linewidth]{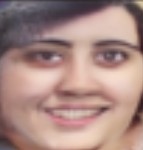}
        \caption{}
        \end{subfigure}\quad
        \begin{subfigure}{0.15\textwidth}
        \centering
        \includegraphics[width=\linewidth]{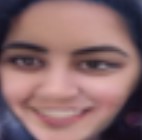}
        \caption{}
        \end{subfigure}\quad
        \begin{subfigure}{0.15\textwidth}
        \centering
        \includegraphics[width=\linewidth]{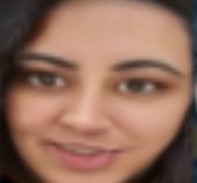}
        \caption{}
        \end{subfigure}\quad
        \caption{Images produced after fine-tuning on StyleGAN}
        \label{fig:2}
    \end{figure} 
    
\begin{table}[h]
  \centering
  \begin{tabular}{l | l l | l }
    \textbf{Image } & \textbf{FN Human User Image 1} & \textbf{FN Human User Image 2} & \textbf{MD}\\
    a & 0.46275 & 0.43165 & 13.63909  \\
    b & 0.60249 & 0.45908 & 13.84692  \\
    c & 0.70137 & 0.69862 & 18.30180 \\
    d & 0.59348 & 0.41899 & 12.89496  \\
    e & 0.46434 & 0.39975 & 10.20973  \\
  \end{tabular}
  \vspace{3mm}
  \caption{Evaluation for Baseline}
  \label{tab:perf}
\end{table}

Images produced using the baseline method (Figure \ref{fig:2}) seem to be less diverse amongst each other and seem to be overfitting on the training dataset. All images except the middle image successfully pass verification. These images appear to have lower diversity (are very similar and look like the same person) and are of relatively lower resolution. 

\subsection{StarGAN v2 Experiments}
\begin{figure}[ht]
        \centering
        \begin{subfigure}{0.15\textwidth}
        \centering
        \includegraphics[width=\linewidth]{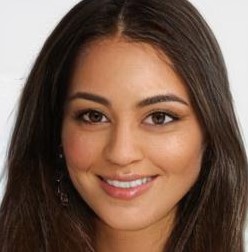}
        \caption{}
        \end{subfigure}\quad
        \begin{subfigure}{0.15\textwidth}
        \centering
        \includegraphics[width=\linewidth]{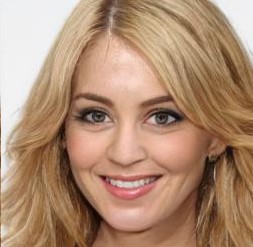}
        \caption{}
        \end{subfigure}\quad
        \begin{subfigure}{0.15\textwidth}
        \centering
        \includegraphics[width=\linewidth]{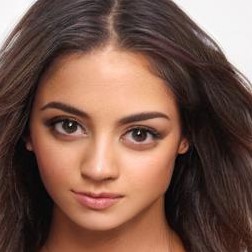}
        \caption{}
        \end{subfigure}\quad
        \begin{subfigure}{0.15\textwidth}
        \centering
        \includegraphics[width=\linewidth]{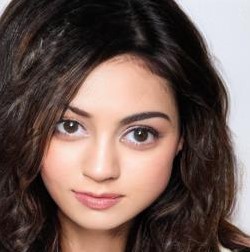}
        \caption{}
        \end{subfigure}\quad
        \begin{subfigure}{0.15\textwidth}
        \centering
        \includegraphics[width=\linewidth]{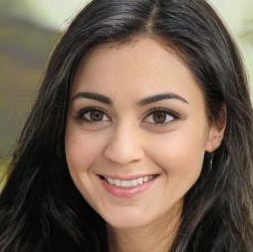}
        \caption{}
        \end{subfigure}\quad
        \begin{subfigure}{0.15\textwidth}
        \centering
        \includegraphics[width=\linewidth]{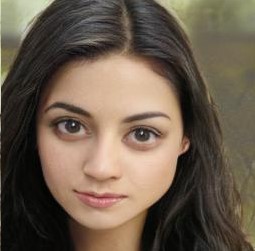}
        \caption{}
        \end{subfigure}\quad
        \begin{subfigure}{0.15\textwidth}
        \centering
        \includegraphics[width=\linewidth]{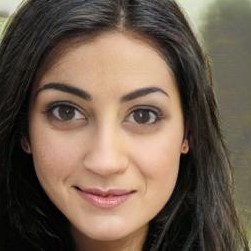}
        \caption{}
        \end{subfigure}\quad
        \begin{subfigure}{0.15\textwidth}
        \centering
        \includegraphics[width=\linewidth]{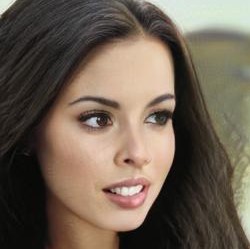}
        \caption{}
        \end{subfigure}\quad
        \caption{Images produced after training on StarGAN v2}
        \label{fig:3}
    \end{figure} 

\begin{table}[h]
  \centering
  \begin{tabular}{l | l l | l }
    \textbf{Image } & \textbf{FN Human User Image 1} & \textbf{FN Human User Image 2} & \textbf{MD}\\
    a & 0.562418 & 0.473044 & 11.098614  \\
    b & 0.632138 & 0.563717 & 14.678981  \\
    c & 0.593215 & 0.431901 & 13.678930 \\
    d & 0.588692 & 0.355675 & 13.722913  \\
    e & 0.473013 & 0.451527 & 12.545832  \\
    f & 0.671785 & 0.516654 & 15.191716  \\
    g & 0.452741 & 0.472176 & 13.059429  \\
    h & 0.716072 & 0.667596 & 21.947438  \\
  \end{tabular}
  \vspace{3mm}
  \caption{Evaluation for StarGAN v2}
  \label{tab:perf}
\end{table}

\begin{figure}[ht]
    \centering
    \includegraphics[width=0.5\textwidth]{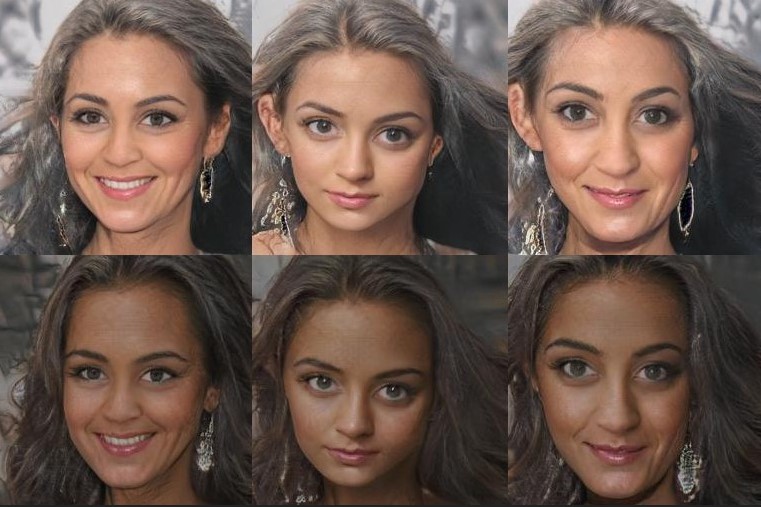}
    \caption{Failed}
    \label{fig:my_label}
\end{figure}

All images (Figure \ref{fig:3}) except the last image successfully pass verification. Images with lower FN distance scores and higher MD scores seem to be appropriate for usage. Images a through d and h used train data as reference and images from the human user dataset as source. Images e through g used images from the human user dataset as both source and reference. Figure \ref{fig:my_label} show sample failures while using images from the human user dataset for both source and reference. This can hopefully be improved after fine tuning on the human user dataset. Another potential experiment is to retrain the model using a combined mixture of the original train data with images from the human user dataset. 

\subsection{Additional StarGAN v2 Experiments}

\subsubsection{Mixed Train Dataset}
\begin{figure}[ht]
    \centering
    \includegraphics[width=0.5\textwidth]{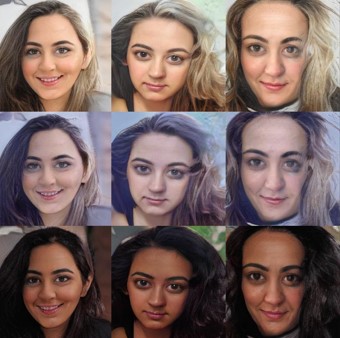}
    \caption{Mixed Train - Failed}
    \label{fig:fail2}
\end{figure}

We attempted training with a mixed dataset consisting of images from both the FairFace dataset and the human user dataset in hopes of producing images that had increased diversity with similar features. However, the model was unable to learn more complex features. So, while the fidelity of the images seem to be better, the images are less diverse and look very alike (Figure \ref{fig:fail2}).

\subsubsection{Freeze-D}
We attempted fine tuning similar to Freeze-D on StarGAN in hopes of producing more accurate high definition images to choose from. However, these images overfit to the train data and ended up being identical to the source image used.

\subsubsection{Gender Reversal}
\begin{figure}[ht]
        \centering
        \begin{subfigure}{0.15\textwidth}
        \centering
        \includegraphics[width=\linewidth]{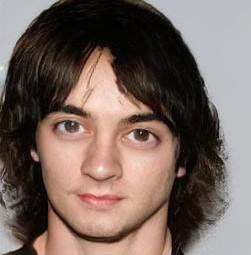}
        \caption{}
        \end{subfigure}\quad
        \begin{subfigure}{0.15\textwidth}
        \centering
        \includegraphics[width=\linewidth]{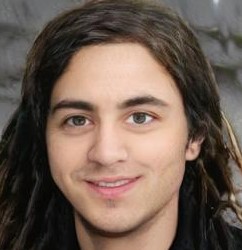}
        \caption{}
        \end{subfigure}\quad
        \begin{subfigure}{0.15\textwidth}
        \centering
        \includegraphics[width=\linewidth]{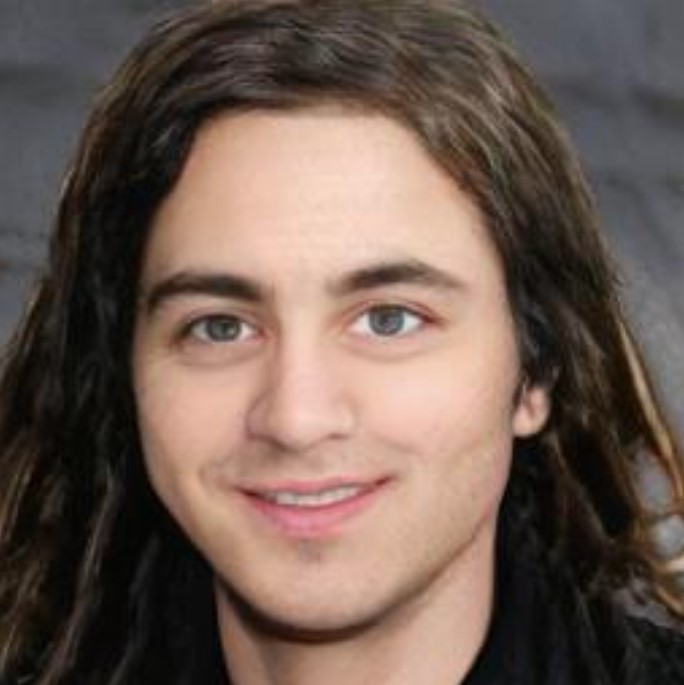}
        \caption{}
        \end{subfigure}\quad
        \caption{Switched Gender Images}
        \label{fig:gr}
    \end{figure}

We also attempted to use our original iteration of StarGAN v2 to generate switched gender images (Figure \ref{fig:gr}). Image c was created from image b. It is interesting to see that image b passes verification while image a and c pass depending on the angle and lighting. It is also interesting to note that these images have a significantly higher  Mahalanobis distance. 
    
\begin{table}[h]
  \centering
  \begin{tabular}{l | l l | l }
    \textbf{Image } & \textbf{FN Human User Image 1} & \textbf{FN Human User Image 2} & \textbf{MD}\\
    a & 0.736005 & 0.617757 & 19.9956032  \\
    b & 0.682950 & 0.596603 & 18.196015  \\
    c & 0.747545 & 0.632504 & 19.905043 \\
  \end{tabular}
  \vspace{3mm}
  \caption{Evaluation for Switched Gender Images}
  \label{tab:perf}
\end{table}

\subsubsection{Dating Application Face Verification Bypass}
\begin{figure}[ht]
        \centering
        \begin{subfigure}{0.37\textwidth}
        \centering
        \includegraphics[width=\linewidth]{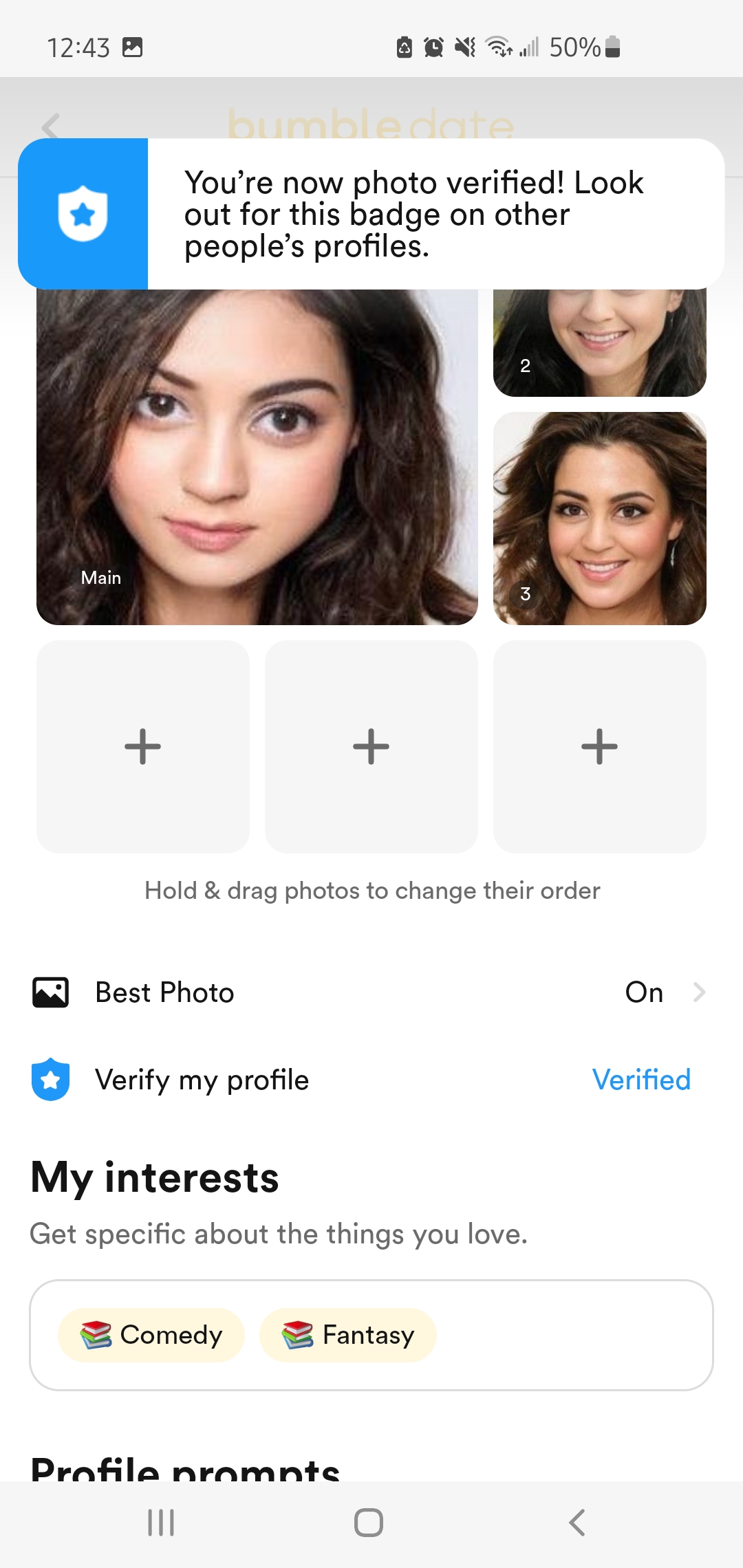}
        \caption{}
        \end{subfigure}\quad
        \begin{subfigure}{0.37\textwidth}
        \centering
        \includegraphics[width=\linewidth]{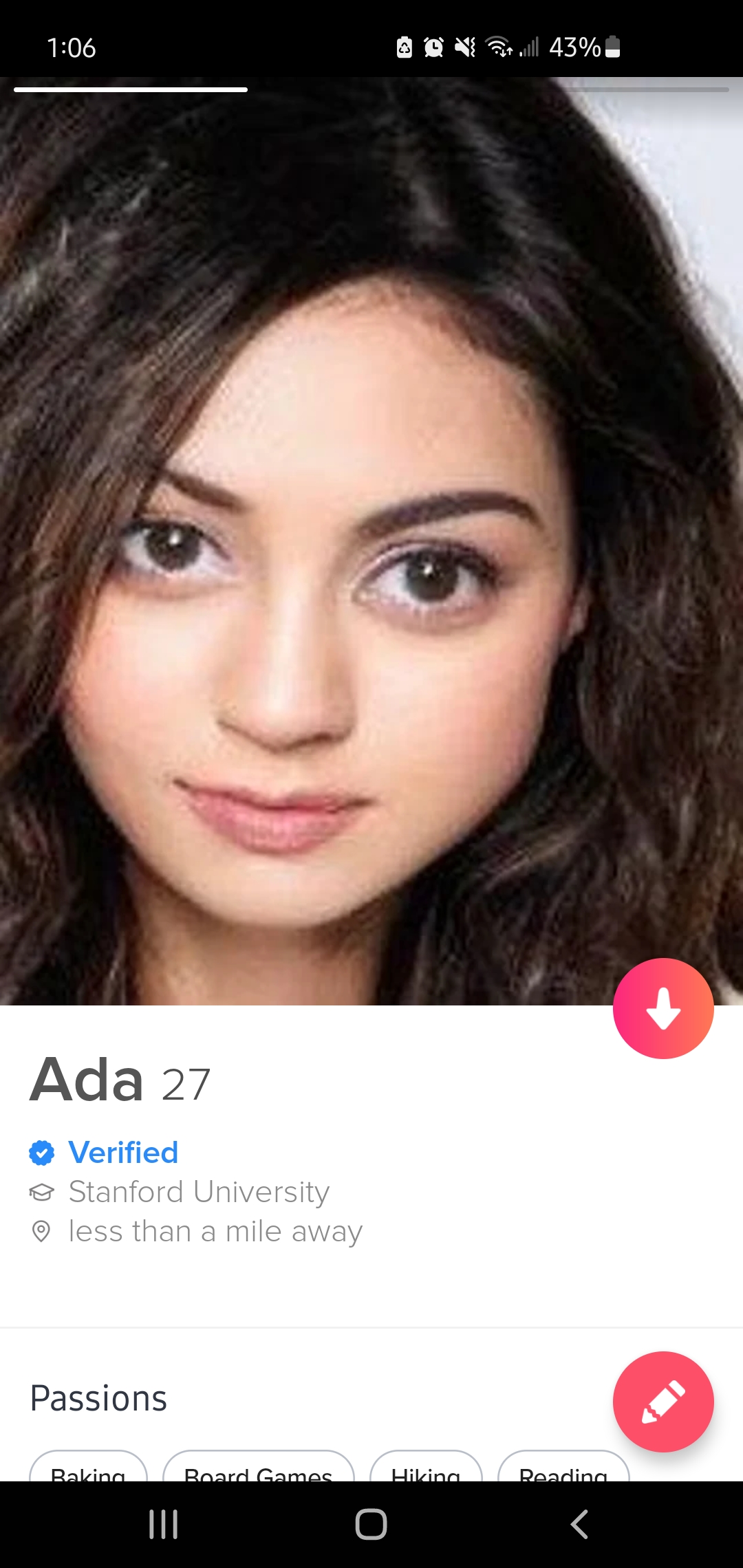}
        \caption{}
        \end{subfigure}\quad
        \begin{subfigure}{0.37\textwidth}
        \centering
        \includegraphics[width=\linewidth]{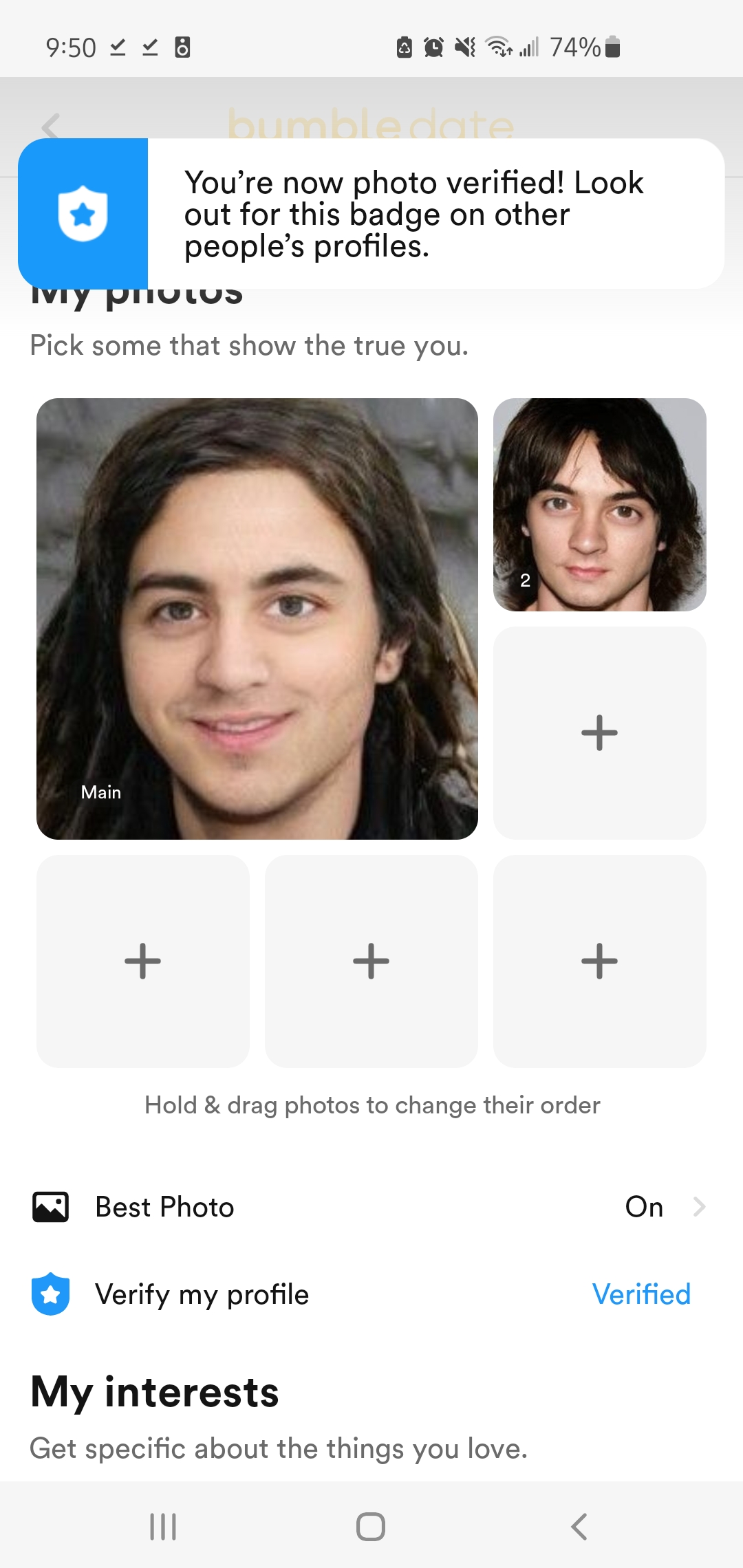}
        \caption{}
        \end{subfigure}\quad
        \caption{Dating App Bypass}
        \label{fig:dab}
    \end{figure}

We tested our generated images on Bumble and Tinder's face verification systems using the human user's actual face (Figure \ref{fig:dab}) and were successfully able to pass face verification. Verifying the switched gender images on Bumble took two attempts with changed lighting conditions. Unfortunately, the switched gender images were unable to pass Tinder's face verification system. Image a and c are from Bumble and image b is from Tinder. 

\section{Conclusion}
In this paper, we show that it is possible to generate images that have similar feature vectors which can be used to bypass face verification systems. We do this by demonstrating a proof of concept on a custom trained white box face verification system and verify our claims by demonstrating the same proof of concept on dating applications that utilize similar face verification systems in a black box setting. We are also able to verify images that are not of the same gender as the human user.

Some areas for future exploration are formalizing this approach using guided search techniques within the feature latent space or potentially diffusion models. Another interesting area would be combining models such as DALL-E \cite{ramesh2021zero} with StarGAN for more controlled image generation. 

Additionally, attempts could be made to reverse the original image from the generated image as a prevention mechanism using techniques such as in \cite{feng2022near}. Of course, more effort should also be made in rigorizing the feature distance metrics currently used in today's face verification systems.

\section*{Acknowledgements}
I would like to thank Anuj Nagpal, Eric Zelikman, and Sharon Zhou for
their valuable discussions and feedback on this work.

\nocite{*}
\bibliographystyle{plain}
\bibliography{references.bib}

\end{document}